\pdfoutput=1

\documentclass[11pt]{article}

\usepackage[]{EMNLP2022}

\usepackage{times}
\usepackage{latexsym}
\usepackage{amsmath}
\usepackage{amssymb}
\usepackage{amsbsy}
\usepackage{enumitem}
\usepackage{subcaption}
\usepackage{multirow}
\usepackage{booktabs}
\usepackage{graphicx}
\usepackage{bbm}
\usepackage{makecell}
\usepackage{tablefootnote}
\usepackage{nccmath}

\usepackage[T1]{fontenc}

\usepackage[utf8]{inputenc}

\usepackage{microtype}

\usepackage{inconsolata}

\def\wv{w}

\def\av{\mathbf{a}}
\def\bv{\mathbf{b}}
\def\cv{\mathbf{c}}
\def\ov{\mathbf{o}}
\def\xv{\mathbf{x}}
\def\xpv{\mathbf{x'}}
\def\yv{\mathbf{y}}

\def\DM{\mathbf{D}}
\def\FM{\mathbf{F}}
\def\KM{\mathbf{K}}
\def\QM{\mathbf{Q}}
\def\VM{\mathbf{V}}

\def\WM{\mathbf{W}}

\def\XPM{\mathbf{X'}}
\def\YM{\mathbf{Y}}

\def\defeq{\overset{\rm def}{=}}

\newcommand{\softmax}[2]{\underset{#1}{\text{\sc SoftMax}}\bigg(#2\bigg)}

\newcommand{\layernorm}[2]{\text{\sc LN}_{#1}(#2)}
\newcommand{\nae}[1]{\text{\sc NAE}_{#1}}
\newcommand{\md}[1]{\text{\sc MD}_{#1}}
\newcommand{\surp}[1]{\mathsf{S}( #1 )}
\newcommand{\condprob}[2]{\mathsf{P}( #1 \mid #2 )}
\newcommand{\ff}[2]{\text{\sc FF}_{#1}(#2)}

\title{Entropy- and Distance-Based Predictors From GPT-2 Attention Patterns Predict Reading Times Over and Above GPT-2 Surprisal}

\author{Byung-Doh Oh \\
  Department of Linguistics \\
  The Ohio State University \\
  \texttt{oh.531@osu.edu} \\\And
  William Schuler \\
  Department of Linguistics \\
  The Ohio State University \\
  \texttt{schuler.77@osu.edu} \\}

\begin{document}
\maketitle
\begin{abstract}
Transformer-based large language models are trained to make predictions about the next word by aggregating representations of previous tokens through their self-attention mechanism.
%
%
In the field of cognitive modeling, such attention patterns have recently been interpreted as embodying the process of cue-based retrieval, in which attention over multiple targets is taken to generate interference and latency during retrieval.
Under this framework, this work first defines an entropy-based predictor that quantifies the diffuseness of self-attention, as well as distance-based predictors that capture the incremental change in attention patterns across timesteps.
Moreover, following recent studies that question the informativeness of attention weights, we also experiment with alternative methods for incorporating vector norms into attention weights.
Regression experiments using predictors calculated from the GPT-2 language model show that these predictors deliver a substantially better fit to held-out self-paced reading and eye-tracking data over a rigorous baseline including GPT-2 surprisal.
%
%
Additionally, the distance-based predictors generally demonstrated higher predictive power, with effect sizes of up to 6.59 ms per standard deviation on self-paced reading times (compared to 2.82 ms for surprisal) and 1.05 ms per standard deviation on eye-gaze durations (compared to 3.81 ms for surprisal).

\end{abstract}


\section{Introduction}
Much work in broad-coverage sentence processing has focused on studying the role of expectation operationalized in the form of surprisal \citep{hale01, levy08} using language models (LMs) to define a conditional probability distribution of a word given its context \citep{smithlevy13, goodkindbicknell18}.
Recently, this has included Transformer-based language models \citep{wilcoxetal20,merkxfrank21,ohetal22}.

\begin{figure}[t!]
    \centering
    \includegraphics[width=0.95\linewidth]{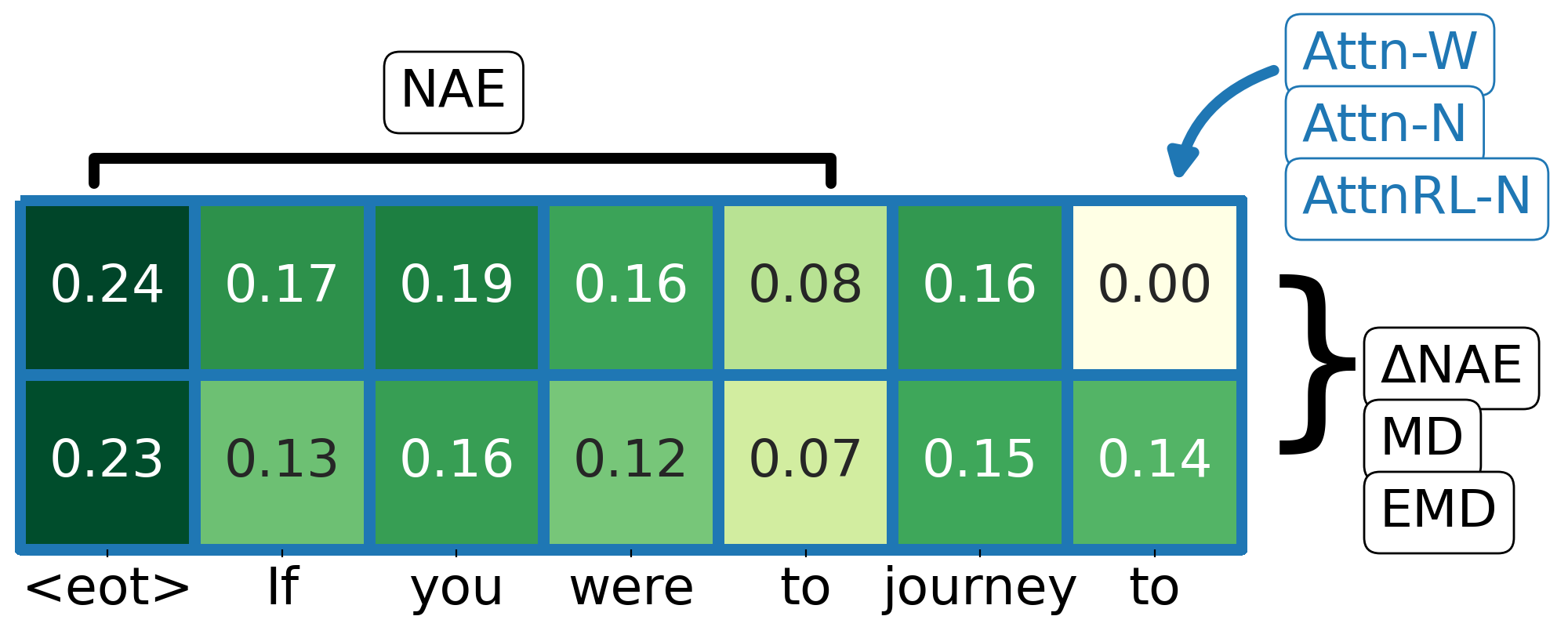}
    \caption{The predictors and attention weight formulations examined in this work. The entropy-based predictor (i.e.~NAE) quantifies the diffuseness of attention over previous tokens at a given timestep, while the distance-based predictors (i.e.~$\Delta$NAE, MD, EMD) capture the change in attention patterns across consecutive timesteps (top row: weights at `journey,' bottom row: weights at `to'). These predictors can be calculated from attention weights formulated using different methods (i.e.~\textsc{Attn-W}, \textsc{Attn-N}, \textsc{AttnRL-N}).}
    \label{fig:overview}
\end{figure}

However, expectation-based accounts have empirical shortcomings, such as being unable to fully account for garden-path effects \citep{vanschijndellinzen21} or predict the timing of delays in certain constructions \citep{levyetal13}.
For this reason, some research has begun to focus on the effects of memory and attention using predictors calculated from language model representations.
%
%
For example, \citet{ryulewis21} recently drew connections between the self-attention patterns of Transformers \citep{vaswanietal17transformer} and cue-based retrieval models of sentence comprehension \citep[e.g.][]{lewisetal06}.
Their attention entropy, which quantifies the diffuseness of the attention weights over previous tokens, showed patterns that are consistent with similarity-based interference observed during the processing of subject-verb agreement.
However, these results relied on identifying one attention head specialized for the \textit{nsubj} dependency, and an aggregated version of this predictor was not very strong in predicting naturalistic reading times in the presence of a surprisal predictor \citep{ryulewis22}.

%
%
This work therefore defines and evaluates several memory- and attention-based predictors derived from the self-attention patterns of the Transformer-based GPT-2 language model \citep{radfordetal19} on two naturalistic datasets, in the presence of a strong GPT-2 surprisal baseline.
First, normalized attention entropy expands upon \citeauthor{ryulewis21}'s \citeyearpar{ryulewis21} attention entropy by re-normalizing the attention weights and controlling for the number of tokens in the previous context.
Additionally, three distance-based predictors that quantify the shift in attention patterns across consecutive timesteps are presented, based on the idea that the reallocation of attentional focus entails processing difficulty.

Moreover, motivated by work on interpreting large language models that question the connection between attention weights and model predictions \citep[e.g.][]{jainwallace19}, the norm-based analysis of the transformed vectors \citep{kobayashietal20, kobayashietal21} is newly applied to GPT-2 in this work to inform alternative formulations of attention weights.
For example, it has been observed that while large language models tend to place high attention weights on special tokens (e.g.~\texttt{<|endoftext|>} of GPT-2), these tokens contribute very little to final model predictions as their `value' vectors have near-zero norms \citep{kobayashietal20}.
Attention weight formulations that incorporate the norms of the transformed vectors should therefore alleviate the over-representation of such special tokens and represent the contribution of each token more accurately.

Results from regression analyses using these predictors show significant and substantial effects in predicting self-paced reading times and eye-gaze durations during naturalistic reading, even in the presence of a robust surprisal predictor.
Additionally, alternative formulations of attention weights that incorporate the norms of the transformed vectors are shown to further improve the predictive power of these predictors.


\section{Background}

This section provides a mathematical definition of the self-attention mechanism underlying the GPT-2 language model and describes alternative norm-based formulations of attention weights.

\subsection{Masked Self-Attention of GPT-2 Language Models}
GPT-2 language models \citep{radfordetal19} use a variant of a multi-layer Transformer decoder proposed in \citet{vaswanietal17transformer}.
Each decoder layer consists of a masked self-attention block and a feedforward neural network:
%
%
%
\begin{equation}
\label{eq:decoder}
\xv_{l+1,i} = \ff{}{\layernorm{\text{out}}{\ov_{l,i} + \xv_{l,i}}} + (\ov_{l,i} + \xv_{l,i}),
\end{equation}
where $\xv_{l,i} \in \mathbb{R}^d$ is the $i$th input representation at layer $l$, \textsc{FF} is a two-layer feedforward neural network, LN$_\text{out}$ is a vector-wise layer normalization operation, and $\ov_{l,i}$ is the output representation from the multi-head self-attention mechanism, in which $H$ multiple heads simultaneously mix representations from the previous context.
This output $\ov_{l,i}$ can be decomposed into the sum of representations resulting from each attention head $h$:
%
%
%
\begin{equation}
\label{eq:attn}
\ov_{l,i} = \sum_{h=1}^{H}{\VM_{h}\begin{bmatrix}\XPM_{l,i} \\ \mathbf{1}^{\top}\end{bmatrix}\av_{l,h,i}},
\end{equation}
where $\XPM_{l,i} \defeq [\xpv_{l,1},...,\xpv_{l,i}] \in \mathbb{R}^{d \times i}$ is the sequence of layer-normalized input representations leading up to $\xpv_{l,i}$ from the previous context, and $\xpv_{l,i} \defeq \layernorm{\text{in}}{\xv_{l,i}}$ is the layer-normalized version of $\xv_{l,i}$.
$\VM_{h}$ represents the head-specific value-output transformation,\footnote{As $\begin{bmatrix} \WM \: \bv \end{bmatrix}$ $\begin{bmatrix} \xv \\ 1 \end{bmatrix} = \WM\xv + \bv$, the bias vectors are omitted from the equations. Additionally, for the simplicity of notation, the `output' affine transform, which applies to the concatenated `value-transformed' vectors from all attention heads in a typical implementation, is subsumed into $\VM_{h}$. The bias vector of the `output' transform is assumed to be distributed equally across heads.} and $\av_{l,h,i} \in \mathbb{R}^{i}$ is the vector of attention weights:
\begin{equation}
\av_{l,h,i} = \softmax{}{\frac{\big(\KM_{h}\!\begin{bmatrix}\XPM_{l,i} \\ \mathbf{1}^{\top}\end{bmatrix}\big)^{\top}\QM_{h}\!\begin{bmatrix}\xpv_{l,i} \\ 1\end{bmatrix}}{\sqrt{d_{h}}}},
\end{equation}
%
where $\QM_{h}$ and $\KM_{h}$ represent the head-specific query and key transformations respectively, and $d_{h} = d/H$ is the dimension of each attention head.
%
%
%
%
%

\begin{figure}[t!]
    \centering
    \includegraphics[width=\linewidth]{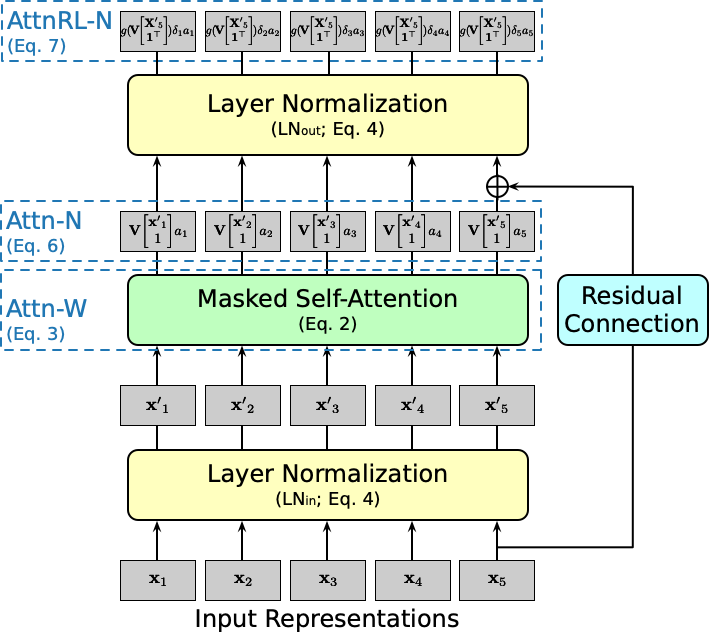}
    \caption{Computations performed within the self-attention block of one head of the GPT-2 language model at a given timestep ($i\!=\!5$). While the masked self-attention mechanism aggregates representations from the previous context in a typical implementation, the linear nature of the subsequent computations allows this aggregation to be deferred to after the residual connection and layer normalization, thereby allowing updated representations to inform alternative formulations of attention weights (i.e.~\textsc{Attn-N}, \textsc{AttnRL-N}).}
    \label{fig:attn_block}
\end{figure}

LN$_\alpha$ is a vector-wise layer normalization operation \citep{baetal16} that first standardizes the vector and subsequently conducts elementwise transformations using learnable parameters $\cv_\alpha, \bv_\alpha \in \mathbb{R}^{d}$:
\begin{equation}
\layernorm{\alpha}{\yv} = \frac{\yv-m(\yv)}{s(\yv)} \odot \cv_\alpha + \bv_\alpha,
\end{equation}
where $\alpha \in \{\text{in}, \text{out}\}$, $m(\yv)$ and $s(\yv)$ denote the elementwise mean and standard deviation respectively, and $\odot$ denotes a Hadamard product.
%

\subsection{Weight- and Norm-Based Analysis of Previous Context}
Previous work that has studied the inner mechanism of Transformer models has focused on analyzing the relative contribution of each token to its final prediction.
As a measure that quantifies the `strength' of contribution, attention weights from the self-attention mechanism have been most commonly used.
Similarly to recent work in cognitive modeling \citep[e.g.][]{ryulewis21}, this work also evaluates predictors calculated from attention weights ({\sc Attn-W}).
%
\begin{equation}
    \av_{\text{W},l,h,i} = \av_{l,h,i}
\end{equation}

While analysis using attention weights is common, the assumption that attention weights alone reflect the contribution of each token disregards the magnitudes of the transformed input vectors, as pointed out by \citet{kobayashietal20}.
As an alternative, they proposed a norm-based analysis of the self-attention mechanism, which quantifies the contribution of each token as the norm of the transformed vector multiplied by the attention weight.
In this work, in order to quantify the relative contribution of each token in the previous context, the norms of the transformed vectors are normalized across the sequence, resulting in `norm-aware' weights that are comparable to attention weights ({\sc Attn-N}).
%
%
\begin{align} \notag
    & \av_{\text{N},l,h,i} = \\ & \medmath{\frac{\bigg(\bigg(\VM_{h}\!\begin{bmatrix}\XPM_{l,i} \\ \mathbf{1}^{\top}\end{bmatrix} \odot \VM_{h}\!\begin{bmatrix}\XPM_{l,i} \\ \mathbf{1}^{\top}\end{bmatrix}\bigg)^{\!\!\top} \mathbf{1}\bigg)^{\!\frac{1}{2}} \!\odot \av_{l,h,i}}{\mathbf{1}^{\top}\bigg(\!\bigg(\!\bigg(\VM_{h}\!\begin{bmatrix}\XPM_{l,i} \\ \mathbf{1}^{\top}\end{bmatrix} \odot \VM_{h}\!\begin{bmatrix}\XPM_{l,i} \\ \mathbf{1}^{\top}\end{bmatrix}\bigg)^{\!\!\top} \mathbf{1}\bigg)^{\!\frac{1}{2}} \!\odot \av_{l,h,i}\bigg)}}  
\end{align}

More recently, \citet{kobayashietal21} showed that the residual connection and the layer normalization operation (\textsc{ResLN; RL}) that follow the self-attention mechanism can also be conducted before aggregating representations over token positions.
Motivated by this observation, the vector norms that take into consideration these subsequent operations are also examined in this work.
Similarly to {\sc Attn-N}, the norms are normalized across the sequence to yield weights that are comparable ({\sc AttnRL-N}):
\begin{align} \notag
    & \av_{\text{RL-N},l,h,i} = \\ & \medmath{\frac{((g(\VM_{h}\!\begin{bmatrix}\XPM_{l,i} \\ \mathbf{1}^{\top}\end{bmatrix}) \odot g(\VM_{h}\!\begin{bmatrix}\XPM_{l,i} \\ \mathbf{1}^{\top}\end{bmatrix}))^{\top} \mathbf{1})^{\frac{1}{2}} \odot \av_{l,h,i}}{\mathbf{1}^{\top}(((g(\VM_{h}\!\begin{bmatrix}\XPM_{l,i} \\ \mathbf{1}^{\top}\end{bmatrix}) \odot g(\VM_{h}\!\begin{bmatrix}\XPM_{l,i} \\ \mathbf{1}^{\top}\end{bmatrix}))^{\top} \mathbf{1})^{\frac{1}{2}} \! \odot \! \av_{l,h,i})}},
\end{align}
\noindent where $g(\cdot)$ incorporates the residual connection ($+\xv_{l,i}$) and layer normalization ($\text{LN}_{\text{out}}$) of Eq.~\ref{eq:decoder}.
Following the assumption that the residual connection serves to `preserve' the representation at position $i$ \citep{kobayashietal21} and that it is distributed equally across heads, $\xv_{l,i}$ is added to the representation at position $i$ of each head after dividing it by the number of heads:
\begin{align} \notag
    & g(\YM) \defeq \\ & \medmath{\footnotesize \sum_{j=1}^{i} \! \left( \!\! \begin{cases} \!
    \frac{\YM\delta_{j} + \frac{\xv_{l,i}}{H\av_{l\!,\!h\!,\!i\![\!i\!]}} - m(\YM\delta_{j} + \frac{\xv_{l,i}}{H\av_{l\!,\!h\!,\!i\![\!i\!]}}\!)}{s(\ov_{l,i}+\xv_{l,i})} \! \odot \! \cv_\text{out} \! + \! \frac{\bv_\text{out}}{H} & \!\!\!\! \text{if } i \!=\! j \\
    \! \frac{\YM\delta_{j} - m(\YM\delta_{j})}{s(\ov_{l,i}+\xv_{l,i})} \! \odot \! \cv_\text{out} \! + \! \frac{\bv_\text{out}}{H} & \!\!\!\! \text{if } i \!\neq\! j
    \end{cases} \!\! \right) \!\! \delta_{j}^{\top}}\!,
\end{align}
\noindent where $\delta_j$ is a Kronecker delta vector consisting of a one at element~$j$ and zeros elsewhere, and $\cv_\text{out}$ and $\bv_\text{out}$ refer to the learnable parameters of $\text{\sc LN}_{\text{out}}$.

\section{Entropy- and Distance-Based Predictors From Attention Patterns}
\label{sec:preds}

Given the different formulations of self-attention weights, entropy-based predictors that quantify the diffuseness of self-attention and distance-based predictors that capture the incremental change in attention patterns across timesteps can be defined.
The first predictor defined in this work is normalized attention entropy (NAE): 
%
\begin{align} \notag 
& \nae{\pi,l,h,i} = \\ & \medmath{\frac{\av_{\pi,l,h,i[1:i-1]}^{\top}}{\log_{2}\!{(i-1)}\mathbf{1}^{\top} \av_{\pi,l,h,i[1:i-1]}} (\log_{2}\!{\frac{\av_{\pi,l,h,i[1:i-1]}}{\mathbf{1}^{\top} \av_{\pi,l,h,i[1:i-1]}}})},
\end{align}
where $\pi \in \{\text{W}, \text{N}, \text{RL-N}\}$.
This is similar to the attention entropy proposed by \citet{ryulewis21} as a measure of interference in cue-based recall attributable to uncertainty about the target, with two notable differences.
First, NAE controls for the number of tokens in the previous context by normalizing the entropy by the maximum entropy that can be achieved at timestep $i$.
Furthermore, NAE also uses weights over $\xpv_{l,1}$, ..., $\xpv_{l,i-1}$ that have been re-normalized to sum to 1, thereby adhering closer to the definition of entropy, in which the mass of interest sums to 1.\footnote{Preliminary analyses showed that regression models with attention entropy proper without these adjustments failed to converge coherently.}

In addition to NAE, distance-based predictors are defined for capturing effortful change in attention patterns across timesteps.
However, as it currently remains theoretically unclear how this distance should be defined, this exploratory work sought to provide empirical results for different distance functions.
The first is $\Delta$NAE, which quantifies the change in diffuseness across timesteps:
\begin{equation}
\Delta\nae{\pi,l,h,i} = |\nae{\pi,l,h,i} \! - \! \nae{\pi,l,h,i-1}|
\end{equation}
As with NAE, this predictor is insensitive to how the attention weights are reallocated between tokens in the previous context to the extent that the overall diffuseness remains unchanged.

The second distance-based predictor is Manhattan distance (MD).\footnote{For the purpose of calculating this predictor, the $i$th element of $\av_{\pi,l,h,i-1}$ is assumed to be 0.}
\begin{equation}
\md{\pi,l,h,i} = ||\av_{\pi,l,h,i}-\av_{\pi,l,h,i-1}||_{1}
\end{equation}
MD directly measures the magnitude of change in attention weights over all tokens at timestep $i$.
MD is less sensitive to the linear distance between tokens and therefore makes it consistent with the predictions of \citet{mcelreeetal03}, who found that processing speed was not influenced by the amount of intervening linguistic material in the formation of a dependency.

Finally, the Earth Mover's Distance \citep[EMD;][]{rubneretal00} is applied to quantify the shift in attention weights.
EMD is derived from a solution to the Monge-Kantorovich problem \citep{rachev85}, which aims to minimize the amount of ``work'' necessary to transform one histogram into another.
Formally, let $P=\{(p_{1},w_{p_{1}}),...,(p_{m},w_{p_{m}})\}$ be the first histogram with $m$ bins, where $p_{r}$ represents the bin and $w_{p_{r}}$ represents the weight of the bin; $Q=\{(q_{1},w_{q_{1}}),...,(q_{n},w_{q_{n}})\}$ the second histogram with $n$ bins; and $\DM=[d_{rs}]$ the distance matrix where $d_{rs}$ is the ground distance between bins $p_{r}$ and $q_{s}$.
The problem is to find an optimal flow $\FM = [f_{rs}]$, where $f_{rs}$ represents the flow between $p_{r}$ and $q_{s}$, that minimizes the overall work.
\begin{equation}
\text{\sc Work}(P, Q, \FM) = \sum_{r=1}^{m}\sum_{s=1}^{n}{d_{rs}f_{rs}}
\end{equation}

Once the optimal flow is found, the EMD is defined as the work normalized by the total flow.\footnote{The optimal flow can be found using the transportation simplex method. Additionally, due to the constraint that the total flow is equal to $\text{min}(\sum_{r=1}^{m}{w_{p_{r}}}, \sum_{s=1}^{n}{w_{q_{s}}}$), the total flow is always equal to 1 in the context of attention weights.} 
\begin{equation}
\text{\sc EMD}(P, Q) = \frac{\sum_{r=1}^{m}\sum_{s=1}^{n}d_{rs}f_{rs}}{\sum_{r=1}^{m}\sum_{s=1}^{n}f_{rs}}
\end{equation}
%
%
To quantify the minimum amount of work necessary to `transform' the attention weights, the EMD between attention weights at consecutive timesteps is calculated using EMD($\av_{\pi,l,h,i-1}$, $\av_{\pi,l,h,i}$).
The ground distance is defined as $d_{rs}=\frac{|r-s|}{i-1}$ in order to control for the number of tokens in the previous context.
EMD can be interpreted as being consistent with Dependency Locality Theory \citep{gibson00} in that reallocating attention weights to tokens further away in the context incurs more cost than reallocating weights to closer tokens.

Code for calculating all of the predictors from GPT-2 under the different attention weight formulations is publicly available at \url{https://github.com/byungdoh/attn_dist}.

%

\section{Experiment 1: Evaluation of Predictors on Human Reading Times}

In order to evaluate the 
contribution 
of the entropy- and distance-based predictors, regression models containing commonly used baseline predictors, surprisal predictors, and one predictor of interest were fitted to self-paced reading times and eye-gaze durations collected during naturalistic language processing.
In this work, we adopt a statistical procedure that directly models temporal diffusion (i.e.~a lingering reponse to stimuli) by estimating continuous impulse response functions and controls for overfitting by assessing the external validity of these predictors through a non-parametric test on held-out data.
%

\subsection{Response Data}

The first experiment described in this paper used the Natural Stories Corpus \citep{futrelletal21}, which contains self-paced reading times from 181 subjects that read 10 naturalistic stories consisting of 10,245 words.
The data were filtered to exclude observations for sentence-initial and sentence-final words, observations from subjects who answered fewer than four comprehension questions correctly, and observations with durations shorter than 100 ms or longer than 3000 ms.
This resulted in a total of 770,102 observations, which were subsequently partitioned into a fit partition of 384,905 observations, an exploratory partition of 192,772 observations, and a held-out partition of 192,425 observations.\footnote{For both datasets, the fit partition, exploratory partition, and held-out partition contain data points whose summed subject and sentence number have modulo four equal to zero or one, two, and three respectively.}
The partitioning allows model selection (e.g.~making decisions about baseline predictors and random effects structure) to be conducted on the exploratory partition and a single hypothesis test to be conducted on the held-out partition, thus obviating the need for multiple trials correction.
All observations were log-transformed prior to regression modeling.

Additionally, the set of go-past durations from the Dundee Corpus \citep{kennedyetal03} also provided the response variable for regression modeling.
The Dundee Corpus contains eye-gaze durations from 10 subjects that read 67 newspaper editorials consisting of 51,501 words.
The data were filtered to remove unfixated words, words following saccades longer than four words, and words at sentence-, screen-, document-, and line-starts and ends.
This resulted in a total of 195,507 observations, which were subsequently partitioned into a fit partition of 98,115 observations, an exploratory partition of 48,598 observations, and a held-out partition of 48,794 observations.
All observations were log-transformed before model fitting.

\subsection{Predictors}

For each dataset, a set of baseline predictors that capture basic, low-level cognitive processing were included in all regression models.
\begin{itemize}[leftmargin=*]
    \setlength\itemsep{0em}
    \item Self-paced reading times \citep{futrelletal21}: word length measured in characters, index of word position within each sentence;
    \item Eye-gaze durations \citep{kennedyetal03}: word length measured in characters, index of word position within each sentence, saccade length, whether or not the previous word was fixated.
\end{itemize}
In addition to the baseline predictors, two surprisal predictors were also included in all regression models evaluated in this experiment.
The first is unigram surprisal as a measure of word frequency, which was calculated using the KenLM toolkit \citep{heafieldetal13} with parameters estimated on the English Gigaword Corpus \citep{parkeretal09}.
The second is surprisal from GPT-2 Small \citep{radfordetal19}, which is trained on $\sim$8B tokens of the WebText dataset.
Surprisal from the smallest GPT-2 model was chosen because it has been shown to be the most predictive of self-paced reading times and eye-gaze durations among surprisal from all variants of GPT-2 \citep{ohetal22}.

Finally, the entropy- and distance-based predictors defined in Section \ref{sec:preds} were calculated from the attention patterns (i.e.~$\av_{\pi,l,h,i}$ where $\pi~\in~\{\text{W}, \text{N}, \text{RL-N}\}$) of heads on the topmost layer of GPT-2 Small.
Contrary to previous studies that analyzed the attention patterns of all layers, this work focuses on analyzing those of the topmost layer out of the concern that the attention patterns of lower layers are less interpretable to the extent that they perform intermediate computations for the upper layers.
Since the topmost layer generates the representation that is used for model prediction, the attention patterns from this layer are assumed to reflect the contribution of each previous token most directly.
%
Subsequently, the by-head predictors were aggregated across heads to calculate by-word predictors.
This assumes that each attention head contributes equally to model prediction, and is also consistent with the formulation of multi-head self-attention in Eq.~\ref{eq:attn}.

\begin{figure*}[ht!]
    \centering
    \includegraphics[width=0.495\linewidth]{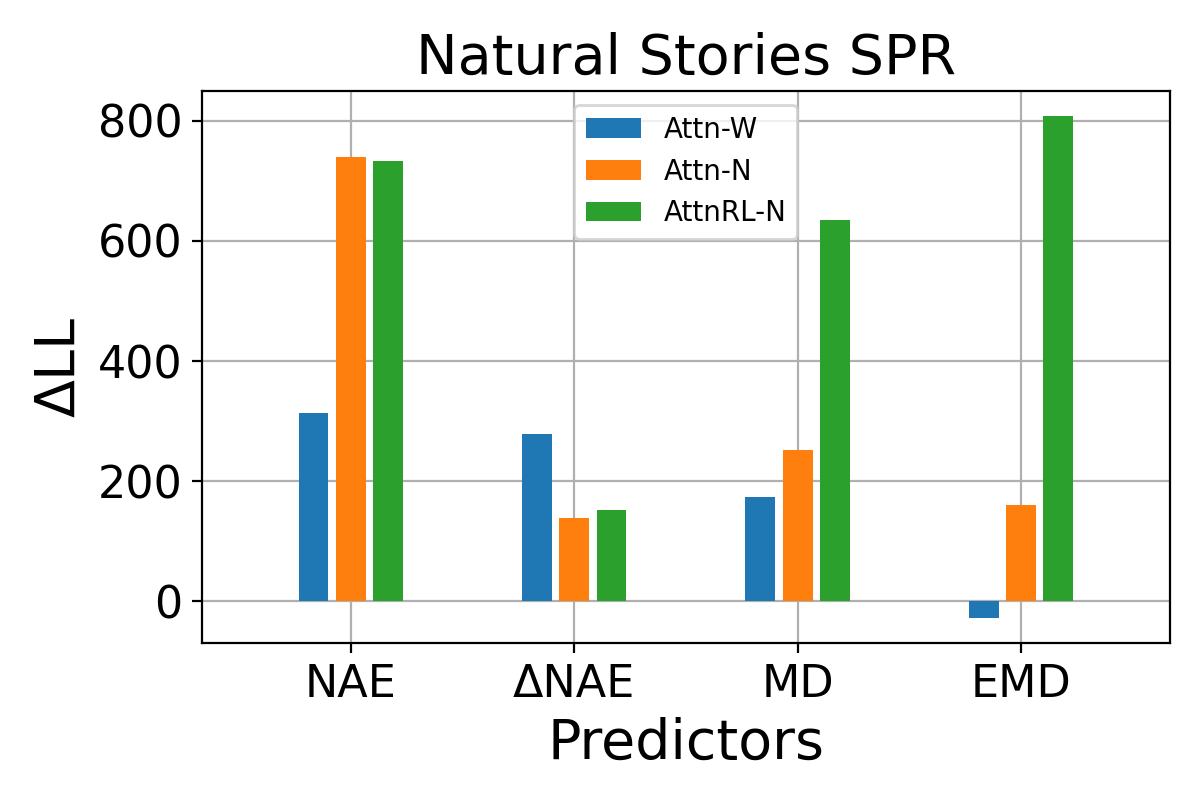}
    \includegraphics[width=0.495\linewidth]{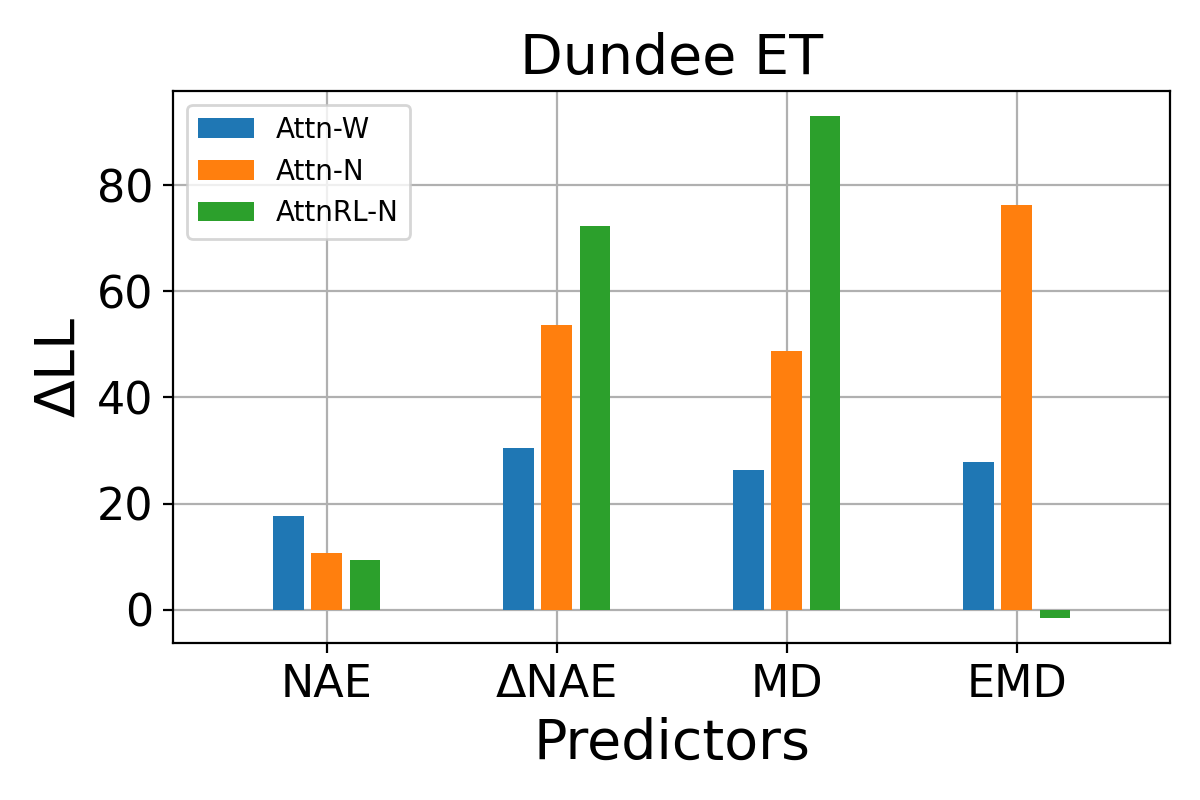}
    \caption{Improvements in CDR model log-likelihood from including each predictor on the exploratory partition of Natural Stories self-paced reading data (left) and Dundee eye-tracking data (right).}
    \label{fig:expl}
\end{figure*}

To calculate surprisal as well as the entropy- and distance-based predictors, each story of the Natural Stories Corpus and each article of the Dundee Corpus was tokenized according GPT-2's byte-pair encoding \citep[BPE;][]{sennrichetal15} tokenizer and was provided as input to the GPT-2 model.
In cases where each story/article did not fit into a single context window, the second half of the previous context window served as the first half of a new context window to calculate predictors for the remaining tokens.\footnote{In practice, most stories/articles fit within two windows.}
Additionally, when a single word $\wv_{i}$ was tokenized into multiple subword tokens, negative log probabilities of subword tokens corresponding to $\wv_{i}$ were added together to calculate $\surp{ \wv_{i} } = -\log\condprob{ \wv_{i} }{ \wv_{1..i-1} }$.
Similarly, the entropy- and distance-based predictors for such subword tokens were also added together.


\subsection{Procedures}

To evaluate the predictive power of each predictor of interest, a {\em baseline} regression model containing only the baseline predictors and {\em full} regression models containing one predictor of interest on top of the baseline regression model were first fitted to the fit partition of each dataset.
In order to control for the confound of temporal diffusion, continuous impulse response functions (IRFs) were estimated using the statistical framework of continuous-time deconvolutional regression \citep[CDR;][]{shainschuler21}.
%
%
The predictors of interest were centered, and all regression models included a by-subject random intercept.\footnote{Early analysis on the exploratory partition revealed that the training data does not support a richer random-effects structure, which led to severe overfitting. Please refer to Appendix \ref{sec:cdr} for details on IRF specifications, the optimization procedure of CDR models, as well as the transformations of baseline predictors.}

As a preliminary analysis, the predictive power of different predictors was compared on the exploratory partition by calculating the increase in log-likelihood ($\Delta$LL) to the baseline regression model as a result of including the predictor, following recent work \citep{goodkindbicknell18, wilcoxetal20, ohetal21acl}.
Subsequently, based on the preliminary exploratory results, the predictive power of one best entropy-based predictor and that of one best distance-based predictor were evaluated on the held-out partition of both datasets.
More specifically, statistical significance testing was conducted using a paired permutation test \citep{demvsar06} of the difference in by-item squared error between the baseline regression model and the respective full regression models.

\subsection{Results}

The results in Figure \ref{fig:expl} show that across both corpora, most of the entropy- and distance-based predictors make a notable contribution to regression model fit under all attention formulations.
Given that the baseline model contains strong predictors such as unigram surprisal and GPT-2 surprisal, this may suggest their validity as predictors of comprehension difficulty.
The exception to this is the EMD predictor, which did not show an increase in likelihood under \textsc{Attn-W} on the Natural Stories Corpus and under \textsc{AttnRL-N} on the Dundee Corpus.
As EMD is sensitive to how the ground distance $d_{rs}$ between bins is defined (i.e.~how costly it is to reallocate attention weights across input positions), a more principled definition of $d_{rs}$ may make EMD a more robust predictor.\footnote{For example, a recency bias may be incorporated by defining the ground distance from tokens closer to the current timestep to be smaller than that from tokens that are farther back in the previous context.}
Across the two corpora, \textsc{Attn-N+NAE} and \textsc{AttnRL-N+MD} appear to be the most predictive among the entropy- and distance-based predictors respectively. 

Additionally, the NAE and $\Delta$NAE predictors showed different trends across the two corpora, where NAE contributed to stronger model fit than $\Delta$NAE on the Natural Stories Corpus, while the opposite trend was observed on the Dundee Corpus.
In contrast to various surprisal predictors that have shown a very similar trend in terms of predictive power across these two corpora \citep{ohetal22}, these two predictors may shed light on the subtle differences between self-paced reading times and eye-gaze durations.
Finally, incorporating vector norms into attention weights (i.e.~\textsc{Attn-N} and \textsc{AttnRL-N}) generally seems to improve the predictive power of these predictors, which provides support for the informativeness of input vectors in analyzing attention patterns \citep{kobayashietal20, kobayashietal21}.

Table \ref{tab:held} presents the effect sizes of \textsc{Attn-N+NAE} and \textsc{AttnRL-N+MD} on the held-out partition of the Natural Stories Corpus and the Dundee Corpus, which were derived by calculating how much increase in reading times the regression model would predict at average predictor value given an increase of one standard deviation.
On both datasets, \textsc{AttnRL-N+MD} appears to be a strong predictor of reading times, which contributed to significantly lower held-out errors.
The entropy-based \textsc{Attn-N+NAE} predictor showed contrasting results across the two corpora, showing a large effect size on the Natural Stories Corpus but not on the Dundee Corpus.
This is consistent with the differential results between NAE and $\Delta$NAE on the exploratory partition of the two corpora and may hint at differences between self-paced reading times and eye-gaze durations.
In terms of magnitude, the two predictors showed large effect sizes on the Natural Stories Corpus, which were more than twice that of GPT-2 surprisal.
On the Dundee Corpus, however, the effect size of \textsc{AttnRL-N+MD} was much smaller compared to that of GPT-2 surprisal.

\begin{table}[t]
    \centering
    \begin{tabular}{c r l}
    \toprule
    Corpus & Predictor & Effect Size \footnotesize{($p$-value)}  \\
    \midrule
    \multirow{2}{*}{\makecell[tc]{Natural \\ Stories}} & {\sc  \footnotesize Attn-N+NAE} & 6.87 ms \footnotesize{($p<0.001$)}  \\
    & {\sc  \footnotesize GPT2Surp} & 2.56 ms \\
    \midrule
    & {\sc \footnotesize AttnRL-N+MD} & 6.59 ms \footnotesize{($p<0.001$)}  \\
    & {\sc \footnotesize GPT2Surp} & 2.82 ms \\
    \midrule
    \multirow{2}{*}{Dundee} & {\sc \footnotesize Attn-N+NAE} & N/A\tablefootnote{The estimated IRF for {\sc Attn-N+NAE} showed a sign error, likely due to poor convergence. Therefore, we treated the effect of this predictor to be statistically non-significant.} \footnotesize{(n.s.)}  \\
    & {\sc  \footnotesize GPT2Surp} & 4.22 ms \\
    \midrule
    & {\sc \footnotesize AttnRL-N+MD} & 1.05 ms \footnotesize{($p<0.001$)}  \\
    & {\sc \footnotesize GPT2Surp} & 3.81 ms \\
    \bottomrule
    \end{tabular}
    \caption{The per standard deviation effect sizes of the predictors on the held-out partition of the Natural Stories Corpus and the Dundee Corpus. Statistical significance was determined by a paired permutation test of the difference in by-item squared error between the baseline regression model and the respective full regression model containing the predictor of interest. The effect sizes of GPT-2 surprisal from the same regression models are presented for comparison.}
    \label{tab:held}
\end{table}

\section{Experiment 2: Do NAE and MD Independently~Explain~Reading~Times?}
The previous experiment revealed that on the Natural Stories Corpus, the select entropy- and distance-based predictors from the attention patterns of GPT-2 contributed to significantly higher regression model fit.
Although they showed similarly large effect sizes on the held-out partition, the two predictors may independently explain reading times, as they are defined to quantify different aspects of attention patterns.
The second experiment examines this possibility following similar procedures as the previous experiment.

\subsection{Procedures}
In order to determine whether the effect of one predictor subsumes that of the other, a CDR model including \textit{both} \textsc{Attn-N+NAE} and \textsc{AttnRL-N+MD} was first fit to self-paced reading times of the fit partition of the Natural Stories Corpus.
The CDR model followed the same specifications, random effects structure, and baseline predictors as described in Experiment 1.
Subsequently, the fit of this regression model on the held-out partition was compared to those of the two regression models that contain only one of the two predictors from the previous experiment.
More specifically, the $\Delta$LL as a result of including the predictor(s) of interest were calculated for each regression model, and statistical significance testing was conducted using a paired permutation test of the difference in by-item squared error between the new `\textsc{Attn-N+NAE} \& \textsc{AttnRL-N+MD}' regression model and the respective `\textsc{Attn-N+NAE}' and `\textsc{AttnRL-N+MD}' regression models, which allowed the contribution of each predictor to be analyzed.

\begin{figure}[t!]
    \centering
    \includegraphics[width=0.95\linewidth]{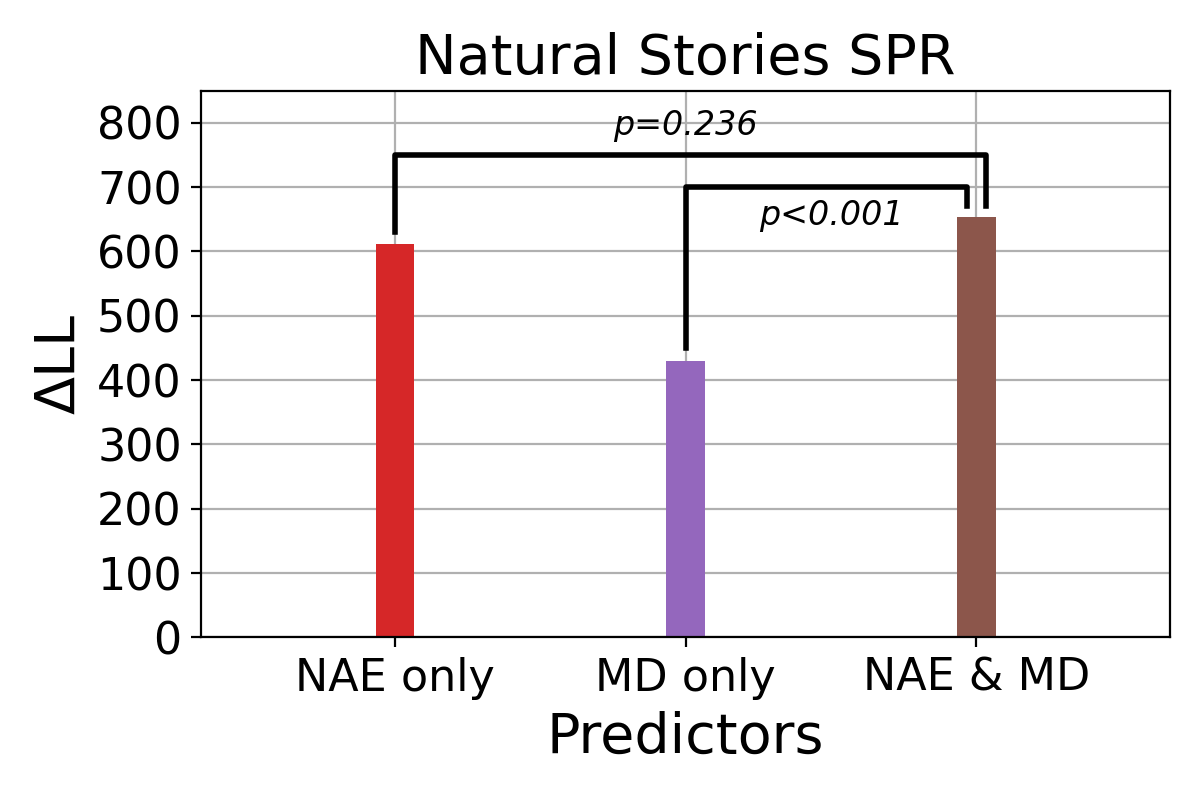}
    \caption{Improvements in CDR model log-likelihood from including \textsc{Attn-N+NAE}, \textsc{AttnRL-N+MD}, and both on the held-out partition of Natural Stories self-paced reading data.}
    \label{fig:held}
\end{figure}

\subsection{Results}

The results in Figure \ref{fig:held} show that the regression model including both \textsc{Attn-N+NAE} and \textsc{AttnRL-N+MD} achieves higher $\Delta$LL on the held-out partition of the Natural Stories Corpus compared to regression models including only one of these predictors.
Moreover, the difference in by-item squared error between the `\textsc{Attn-N+NAE} \& \textsc{AttnRL-N+MD}' regression model and the `\textsc{AttnRL-N+MD}' regression model was statistically significant ($p<0.001$).
In contrast, the significance between the `\textsc{Attn-N+NAE} \& \textsc{AttnRL-N+MD}' regression model and the `\textsc{Attn-N+NAE}' regression model was not statistically significant ($p=0.236$).
This indicates that  the entropy-based \textsc{Attn-N+NAE} contributes to model fit over and above the distance-based \textsc{AttnRL-N+MD} and also subsumes its effect in predicting reading times.

\begin{figure*}[ht!]
    \centering
    \includegraphics[width=0.9\linewidth]{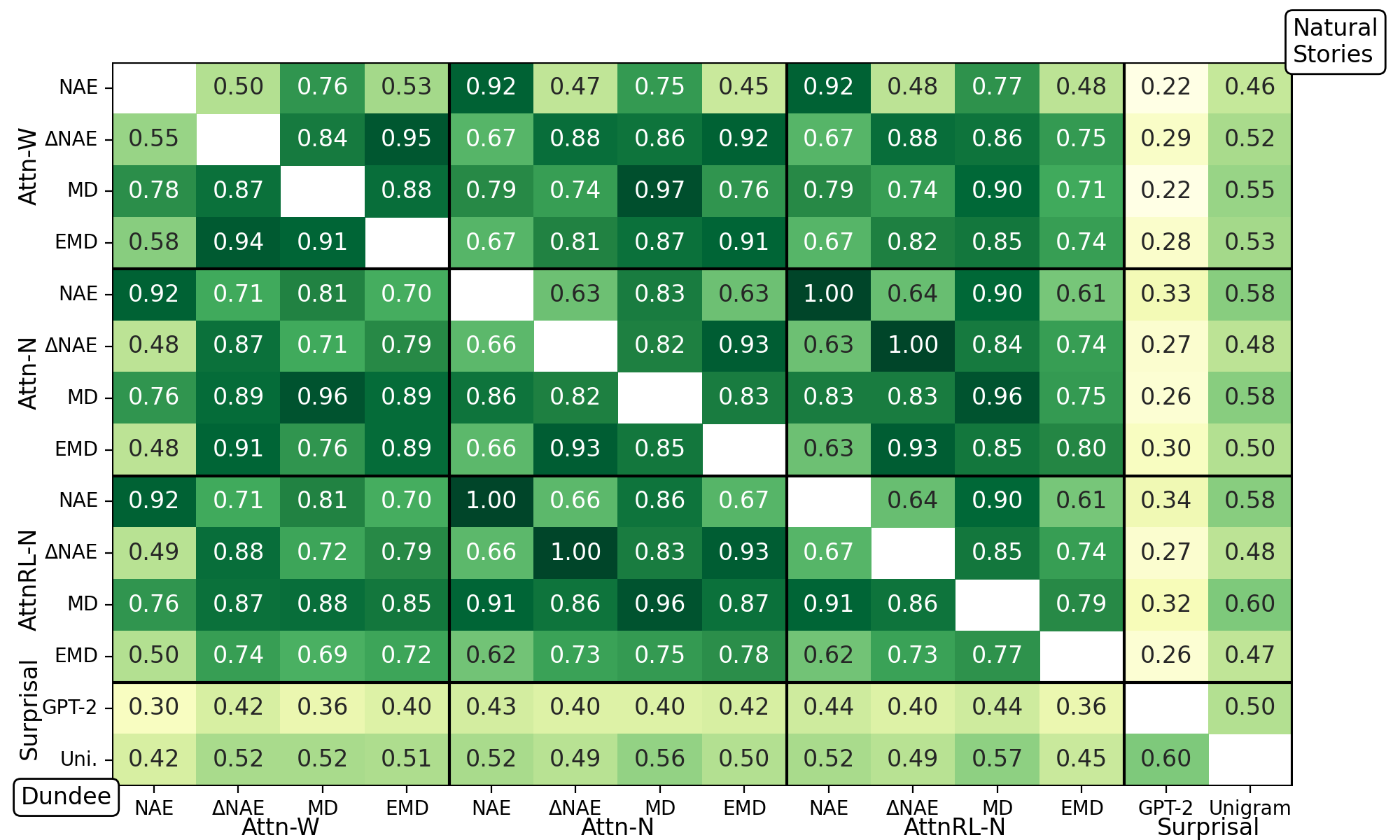}
    \caption{Pearson correlation coefficients between the predictors examined in this work on the Natural Stories Corpus (upper right triangle) and the Dundee Corpus (lower left triangle).}
    \label{fig:corr}
\end{figure*}

\section{Correlation Analysis Between Predictors and Syntactic Categories}

\subsection{Procedures}

In order to shed more light on the predictors newly proposed in this work, a series of correlation analyses were conducted.
First, Pearson correlation coefficients were calculated between the entropy- and distance-based predictors as well as the surprisal predictors to examine the similarity between predictors and the influence of different attention weight formulations.
Subsequently, the most predictive \textsc{Attn-N+NAE} and \textsc{AttnRL-N+MD} predictors were analyzed with a focus on identifying potential linguistic correlates.
This analysis used a version of the Natural Stories Corpus and the Dundee Corpus that had been manually annotated using a generalized categorial grammar annotation scheme \citep{shainetal18:lincr}.

\subsection{Results}
The correlation matrix in Figure \ref{fig:corr} shows that within the same predictor, the different attention formulations did not make a very large difference, with `intra-predictor' correlation coefficients at above $0.85$ for most predictors.
An exception to this trend was EMD, which showed a correlation coefficient of $0.74$ on the Natural Stories Corpus and $0.72$ on the Dundee Corpus between \textsc{Attn-W+EMD} and \textsc{AttnRL-N+EMD}.
Such difference is to be expected, as EMD is the most sensitive to the change of location in attention weights among the distance-based predictors.
This is also consistent with the exploratory regression results in Figure \ref{fig:expl}, where the $\Delta$LL of EMD predictors varied the most as a function of different attention weight formulations.

Additionally, the norm-based attention formulations seem to bring the entropy-based NAE closer to the distance-based predictors in terms of correlation coefficients.
On both corpora, \textsc{Attn-N+NAE} and \textsc{AttnRL-N+NAE} show stronger correlations to distance-based predictors than \textsc{Attn-W+NAE}.
Interestingly, the highest correlation coefficient between NAE and a distance-based predictor is observed between \textsc{Attn-N+NAE} and \textsc{AttnRL-N+MD} at $0.90$ on the Natural Stories Corpus and $0.91$ on the Dundee Corpus, which were the two strongest predictors identified in Experiment 1.
Such high correlation partially explains the results of Experiment 2, in which the influence of \textsc{Attn-N+NAE} subsumed that of \textsc{AttnRL-N+MD}.

Finally, the entropy- and distance-based predictors showed moderate correlation to unigram surprisal at around $0.5$ on both corpora.
With regard to GPT-2 surprisal, these predictors showed weak correlation at around $0.3$ on the Natural Stories Corpus, and around $0.4$ on the Dundee Corpus.
Together with the regression results, this further suggests that the proposed predictors capture a mechanistic process that is distinct from the frequency or predictability of the word.

An analysis of the predictors according to syntactic categories showed that \textsc{AttnRL-N+MD} may be sensitive to the transition between the subject constituent and the predicate constituent.
Its average values for nouns in different contexts presented in Table \ref{tab:opaa} show that on both corpora, nouns occurring at the end of subject NPs were associated with a greater shift in attention patterns.
This is consistent with the intuition that transitions between constituents entails cognitive effort during incremental processing.

\begin{table}[t]
    \centering
    \begin{tabular}{c r r r}
    \toprule
    Corpus & Nouns & Count & Mean  \\
    \midrule
    \multirow{2}{*}{\makecell[tc]{Natural \\ Stories}} & End of subj.~NP & 1,051 & 19.024 \\
    & Other & 3,098 & 16.520 \\
    \midrule
    \multirow{2}{*}{Dundee} & End of subj.~NP & 4,496 & 18.982 \\
    & Other & 15,763 & 16.661 \\
    \bottomrule
    \end{tabular}
    \caption{Average \textsc{AttnRL-N+MD} values for nouns at the end of subject NPs and nouns in other contexts.}
    \label{tab:opaa}
\end{table}

\section{Discussion and Conclusion}

This work builds on recent efforts to derive memory- and attention-based predictors of processing difficulty to complement expectation-based accounts of sentence processing.
Based on the observation that the self-attention patterns of Transformer-based language models can be interpreted as embodying the process of cue-based retrieval, an entropy-based predictor that quantifies the diffuseness of self-attention was first defined.
Moreover, based on the idea that reallocation of attention may incur processing difficulty, distance-based predictors that capture the incremental change in attention patterns across timesteps were defined.
Regression results using these predictors calculated from the GPT-2 language model showed that these entropy- and distance-based predictors deliver a substantially better fit to self-paced reading and eye-tracking data over a strong baseline including GPT-2 surprisal.

To our knowledge, this work is the first to report robust effects of Transformer attention-based predictors in predicting reading times of broad-coverage naturalistic data.
This provides support for \citeauthor{ryulewis21}'s \citeyearpar{ryulewis21} observation that the self-attention mechanism of Transformers embodies the process of cue-based retrieval, and further suggests that representations that exhibit similarity-based interference can be learned from the self-supervised next-word prediction task.
Additionally, the strength of the distance-based predictors further demonstrates the potential to bring together expectation- and memory-based theories of sentence processing under a coherent framework.

\section*{Acknowledgments}
We thank the reviewers for their helpful comments.
This work was supported by the National Science Foundation grant \#1816891.
All views expressed are those of the authors and do not necessarily reflect the views of the National Science Foundation.

\section*{Limitations}
The connection between attention patterns of Transformer-based language models and human sentence processing drawn in this work is based on a model trained on English text and data from human subjects that are native speakers of English.
Therefore, the connection made in this work may not generalize to other languages.
Additionally, although the alternative formulations of self-attention weights resulted in stronger predictors of processing difficulty, they are more computationally expensive to calculate as they rely on an explicit decomposition of the matrix multiplication operation, which are highly optimized in most packages.





\section*{Ethics Statement}
Experiments presented in this work used datasets from previously published research \citep{futrelletal21, kennedyetal03}, in which the procedures for data collection and validation are outlined.
As this work focuses studying the possible connection between attention patterns of large language models and human sentence processing, its potential negative impacts on society seem to be minimal.

\bibliography{emnlp2022}
\bibliographystyle{acl_natbib}

\appendix
\section{CDR Implementation Details}
\label{sec:cdr}
The continuous-time deconvolutional regression (CDR) models used in this work were fitted using variational inference to estimate the means and variances of independent normal posterior distributions over all model parameters assuming an improper uniform prior.
Convolved predictors used the three-parameter ShiftedGamma impuse response function (IRF) kernel:
\begin{equation}
f(x;\alpha,\beta,\delta) = \frac{\beta^{\alpha}(x-\delta)^{\alpha-1}e^{-\beta(x-\delta)}}{\Gamma(\alpha)}
\end{equation}

Posterior means for the IRF parameters were initialized at $\alpha = 0.2$, $\beta = 0.5$, and $\delta = -0.2$, which defines a decreasing IRF with a peak centered at $t=0$ that decays to near-zero within about 1 s. Models were fitted using the Adam optimizer \citep{kingmaba15} with Nesterov momentum \citep{nesterov83, dozat16}, a constant learning rate of 0.001, and minibatches of size 1,024. For computational efficiency, histories were truncated at 128 timesteps. 
Prediction from the network used an exponential moving average of parameter iterates with a decay rate of 0.999, and the models were evaluated using \textit{maximum a posteriori} estimates obtained by setting all IRF parameters to their posterior means.

For the baseline regression predictors, the `index of word position within each sentence' predictors were scaled, and the `word length in characters' and `saccade length' predictors were both centered and scaled.

\end{document}